\documentclass{article}

\usepackage{microtype}
\usepackage{graphicx}
\usepackage{subfigure}
\usepackage{booktabs}
\usepackage{hyperref}

\usepackage[accepted]{conference_style}
\usepackage{amsmath}
\usepackage{amssymb}
\usepackage{mathtools}
\usepackage{amsthm}
\usepackage[capitalize,noabbrev]{cleveref}
\usepackage{array}
\usepackage{ragged2e}
\usepackage{adjustbox}
\usepackage{multirow}
\usepackage{makecell}
\usepackage{textcomp}
\usepackage{placeins}
\usepackage[eulergreek]{sansmath}
\usepackage{pgfplots,pgfplotstable}
\usepackage{cellspace}
\usepackage{tabularx}
\usepackage{authblk}
\usepackage{hyperref}
\usepackage{setspace}
\pgfplotsset{compat=1.18}

\definecolor{myred}{RGB}{183,0,67}
\definecolor{myblue}{RGB}{29,62,228}
\definecolor{myellow}{RGB}{234,173,0}
\definecolor{mygreen}{RGB}{20,150,62}

\theoremstyle{plain}

\theoremstyle{definition}

\theoremstyle{remark}

\icmltitlerunning{Classifying Nodes in Graphs without GNNs}

\newcommand{\modelname}{CoHOp\,}
\newcommand{\cora}{\texttt{Cora}}
\newcommand{\citseer}{\texttt{Citeseer}}
\newcommand{\pubmed}{\texttt{Pubmed}}
\newcommand{\acomputer}{\texttt{A-computer}}
\newcommand{\aphotos}{\texttt{A-photo}}
\newcommand{\ogbp}{\texttt{Products}}
\newcommand{\ogba}{\texttt{Arxiv}}

\newcommand{\nosmog}{\textsf{\small NOSMOG}}
\newcommand{\seen}{\textit{seen}}
\newcommand{\unseen}{\textit{unseen}}
\newcommand{\production}{\textit{prod}}

\begin{document}

\twocolumn[
\icmltitle{Classifying Nodes in Graphs without GNNs}

\begin{center} 
\textbf{Daniel Winter \hspace{1em} Niv Cohen \hspace{1em} Yedid Hoshen}
\vspace{1em}

\texttt{\{daniel.winter, niv.cohen2, yedid.hoshen\}@mail.huji.ac.il}
\vspace{1em}

School of Computer Science and Engineering The Hebrew University of Jerusalem, Israel
\end{center}

\vskip 0.3in
]

\begin{abstract}
Graph neural networks (GNNs) are the dominant paradigm for classifying nodes in a graph, but they have several undesirable attributes stemming from their message passing architecture. Recently, distillation methods succeeded in eliminating the use of GNNs at test time but they still require them during training. We perform a careful analysis of the role that GNNs play in distillation methods. This analysis leads us to propose a fully GNN-free approach for node classification, not requiring them at train or test time. Our method consists of three key components: smoothness constraints, pseudo-labeling iterations and neighborhood-label histograms. Our final approach can match the state-of-the-art accuracy on standard popular benchmarks such as citation and co-purchase networks, without training a GNN\footnote{The full source code of our method is available at \href{https://github.com/dani3lwinter/CoHOp}{https://github.com/dani3lwinter/CoHOp}}.
\end{abstract}

\section{Introduction}

Node classification tasks naturally occur when we wish to classify graph-structured data, such as paper citation networks \cite{coraciteseer, pubmed} or product co-purchase networks \cite{amazondatasets}. Most state-of-the-art node classification methods use Graph Neural Networks (GNN) \cite{gcn,sage,gat}. GNNs exploit the context of the entire neighborhood of each node for determining its class. Their architecture uses message passing to transfer information across many nodes. GNNs consider large neighborhoods, while this is very effective, their training and inference times are considerably higher than methods considering only the node features. 

Therefore, several research efforts attempted to find simpler alternatives to GNNs that do not require message passing. Distillation methods, starting with the seminal work by Zhang et al. \cite{glnncite}, proposed to replace GNNs at test time by simple node-level MLPs. They first train a GNN on the training data, then use the labels predicted by the GNN as distillation targets for training a node-level MLP. At test time, these methods only use the node-level MLP. Distillation methods sometimes achieve even better results than the original GNN. However, while this line of work removed the requirement for GNNs at \textit{test-time}, it still requires \textit{training} a GNN for producing the distillation targets. 

Here, we aim to take this research effort one step further and create an entirely \textit{GNN-free} node classification approach. We begin by investigating the reasons for the success of distillation methods. We suggest the ability to achieve state-the of-the-art performance without using GNNs at test-time hints that their advantage lies in GNNs'  sample efficiency rather than message passing. 

Our findings motivate us to propose CoHOp, a new node classification method consisting of the following key components: (i) A loss encouraging smoothness between the label predictions of neighboring nodes (ii) Iterative pseudo-labelling of the observed unlabelled nodes (iii) Label neighborhood-histogram for encoding local context. CoHOp does not require training or evaluating GNNs at all, and achieves competitive results with GNNs on popular datasets such as citation and co-purchase networks.

Our key contributions are:
\begin{itemize}
    \item Clarifying the role of distillation methods in node classification.
    \item Introducing neighborhood-histogram features to incorporate local context information.
    \item Achieving competitive results on popular node-classification datasets without training graph neural networks.
\end{itemize}

\section{Related works}

\textbf{Graph Neural Networks (GNNs)} have emerged as a prominent tool in the domain of graph machine learning \cite{bruna2013spectral, defferrard2016convolutional, li2019deepgcns, chen2020simple}. These neural networks use aggregations of features from the local context of each node at successive layers. For example, Graph Convolutional Networks (\textit{GCN}) \cite{gcn} extend traditional convolution operations from the Euclidean domain to operations on graphs. \textit{GraphSAGE} \cite{sage} uses arbitrary aggregation function while also concatenating the features prior to the aggregation. GAT \cite{gat}, GTN \cite{yun2019graph} and HAN \cite{wang2019heterogeneous} methods generalize attention layers and transformers to graphs. Many GNNs are formulated into a unified framework called \textit{Message Passing} Neural Networks \cite{mpnn}. 

\textbf{Knowledge Distillation of GNNs.}
Addressing challenges related to memory consumption and latency, several methods have been proposed to distill knowledge from a large pre-trained GNN teacher model to a smaller student model. The student model can be either a smaller GNN model \cite{lee2019graph, yang2020distilling, yan2020tinygnn, tian2023knowledge, guo2023boosting}, or structure-agnostic model.
One such method, \textit{GLNN} \cite{glnncite}, trains MLP model to predict soft-labels obtained from pre-trained GNN. Another approach, \textit{NOSMOG} \cite{nosmogcite}, uses the same underlying method with the addition of adversarial feature augmentation loss and \textit{Similarity Distillation} of hidden features. NOSMOG also utilizes the graph structure by concatenating positional features obtained using DeepWalk \cite{deepwalk}. While NOSMOG offers better accuracy results than standard GLNN, it suffers from higher latency induced by positional feature computation. \textit{CPF} \cite{yang2021extract} also uses a non-GNN student model, although the student still relies on iterative label propagation during inference time, which increases the inference running time.

\textbf{Node classification without GNN.}
Various techniques beyond Graph Neural Networks have been developed. Among them are \textit{Graph-MLP} \cite{graphMLP} which trains an MLP model with a neighbor contrastive loss. Another method, more similar to our approach, \textit{Correct and Smooth} (C\&S) \cite{cands} also leverages the correlation between neighbors' labels to enhance a linear or a shallow MLP predictor. However, it deviates from our approach by refining predictions post-training through label propagation. Moreover, the applicability of the C\&S method to the inductive case (where new nodes are added to the graph during test time) is limited, and it focuses on the supervised, rather than semi-supervised settings (see Sec. \ref{sec:PseudoLabelling}). Also, in contrast to the heavy reliance on labels of C\&S, a significant aspect of our approach addresses the challenges arising from a small training set in semi-supervised scenarios. 

\textbf{Semi-Supervised Learning.} SSL is an approach for leveraging unlabeled data, often used in scenarios where the size of the training set is small. A popular SSL method, termed \textit{pseudo-labeling}, uses the model's predictions as labels for training \cite{mclachlan1975iterative, rosenberg2005semi, lee2013pseudo, xie2020self}. Another prominent SSL approach is consistency regularization \cite{bachman2014learning, sajjadi2016regularization, laine2016temporal}, where the model is enforced to maintain consistent predictions through random augmentation of its input. FixMatch \cite{sohn2020fixmatch} combines these ideas in a simple manner.

\section{Motivation}
\label{sec:motivation}

\begin{figure}
  \centering
  \begin{tikzpicture}
    \begin{axis}[
      xlabel={Portion of the samples in the training set},
      ylabel={Accuracy},
      legend pos=south east,
      legend style={font=\fontsize{8}{8}\selectfont},
      grid=major,
      ymin=65,
      ymax=95,
      height=5.5cm,
      width=\linewidth,
      xticklabel={$\pgfmathprintnumber{\tick}\%$},
      yticklabel={$\pgfmathprintnumber{\tick}\%$},
      xticklabel style={font=\scriptsize\sansmath\sffamily},
      yticklabel style={font=\scriptsize\sansmath\sffamily},
      ]

    \addplot[color=myblue,mark=triangle*, smooth, mark size=1pt, line width=1pt] coordinates {
      (0.5, 79.2)
      (1, 81.5)
      (2.5, 84.0)
      (5, 85.3)
      (10, 86.5)
      (20, 87.6)
      (40, 87.9)
      (60, 88.0)
      (80, 88.0)
    };
    
    \addplot[color=mygreen,mark=*, smooth, mark size=1pt, line width=1pt] coordinates {
      (0.5, 80.4)
      (1, 82.8)
      (2.5, 85.4)
      (5, 86.7)
      (10, 87.8)
      (20, 88.6)
      (40, 89.0)
      (60, 89.2)
      (80, 89.3)
    };

    \addplot[color=myred,mark=square*, smooth, mark size=1pt, line width=1pt] coordinates {
      (0.5, 70.02)
      (1, 73.36)
      (2.5, 77.26)
      (5, 79.81)
      (10, 81.54)
      (20, 83.16)
      (40, 85.67)
      (60, 87.03)
      (80, 87.51)
    };

    \legend{GNN, Distillation, No Distillation}
    \end{axis}
  \end{tikzpicture}
  \vspace{-10pt}
  \caption{ 
   A linear model using node-only features achieves comparable performance to a full GNN on the PubMed dataset. However, its performance degrades quickly as the training set size decreases.
    }
  \label{fig:accuracy_vs_labelrate}
\end{figure}
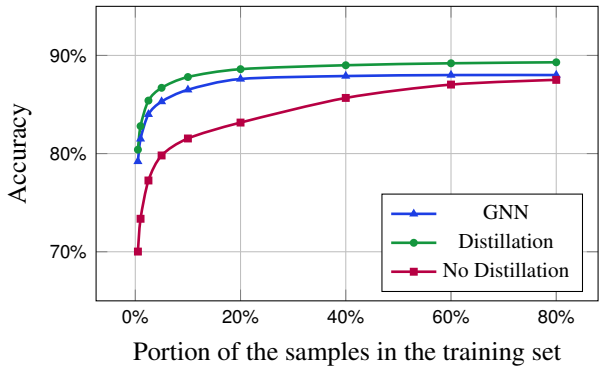

\begin{figure*}[!t]
  \centering
  \includegraphics[width=\textwidth]{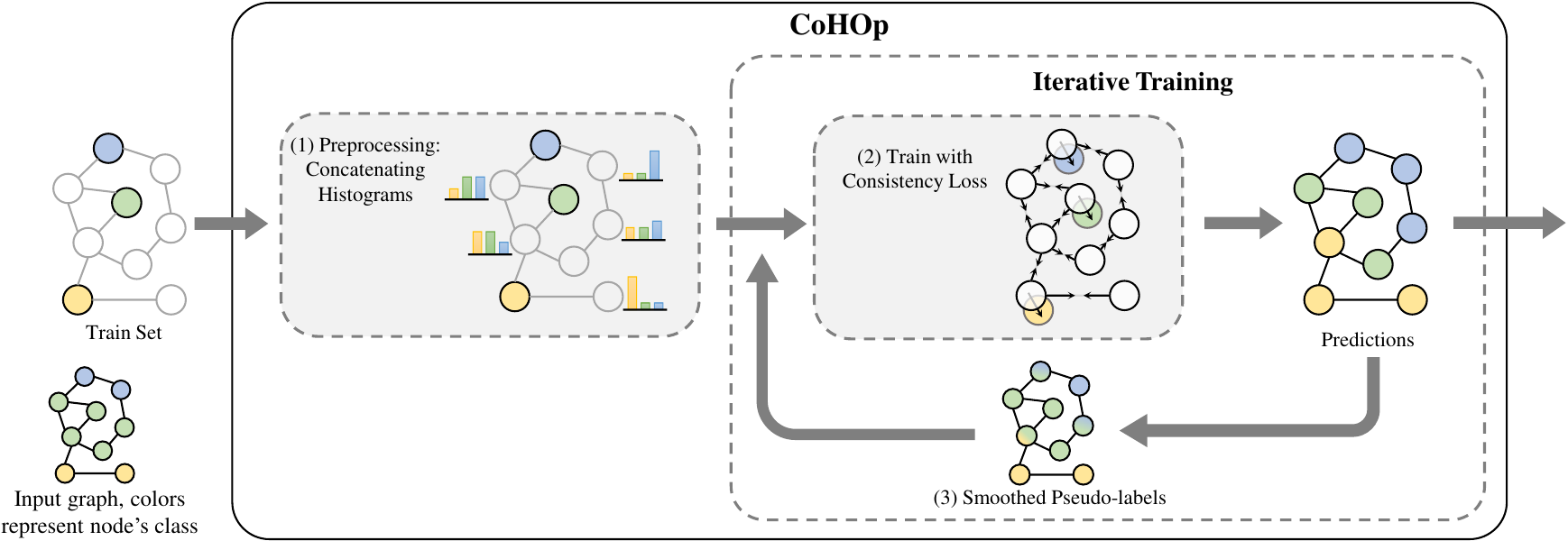}
  \vskip -0.1in
  \caption{Overview of \modelname. Our method consists of three elements: (1) augmenting the node features by concatenating them with the histogram of nearby node labels (Sec. \ref{sec:histograms}). (2) training with consistency loss in addition to the standard cross-entropy classification loss (Sec \ref{sec:consistency}). (3) iterative training with smoothed pseudo-labels (Sec. \ref{sec:PseudoLabelling}). }
  \label{fig:diagram}
  \vskip -0.1in
\end{figure*}

Distillation methods have recently challenged the existing paradigm in node classification. The standard practice with GNNs is to train the model on all the labeled nodes in the graph and use the same model for node classification at test time. Distillation methods remove the need for using a GNN at test time, although they still require training a GNN. They use the GNN for pseudo-labeling unsupervised nodes. Subsequently, an MLP is trained to predict the labels of these nodes based on its node features only, without considering the features of its neighbors. Remarkably, distillation methods are competitive with GNNs on popular citation and co-purchase network benchmarks. This result is confusing, as the graph structure appears beneficial during training but not during test time. This begs the question: \textit{Why are distillation methods so successful?}

To address this question, we examine whether the power of the GNN in this case comes from its increased expressiveness attributed to message passing or rather from its useful inductive bias. We plot the node classification accuracy of both GNNs and node-level MLPs as a function of the training set size (Fig. \ref{fig:accuracy_vs_labelrate}). The observed trend indicates that with an increase in training size, the performance gap between node-level MLPs and GNNs diminishes. This suggests that MLPs overfit due to small training sizes on popular node classification datasets; while GNNs are implicitly better regularized (i.e., they have a useful inductive bias). 

Next, we examine the gap between GNNs and MLPs trained on the distillation targets. Here, we observe that the gap between the two models is narrow, even for small labeled training sets. These experimental findings lead us to conclude that: \textit{the challenge in the examined dataset lies not in increasing model expressivity, but rather in decreasing model sample complexity}. GNNs overcome this challenge through a useful inductive bias, while distillation overcomes it by increasing the size of the training set with GNN pseudo-labels.

In light of this finding, we ask: \textit{are there alternative methods for overcoming overfitting beyond training a full GNN?} The answer to this question is affirmative as will be shown in Sec.~\ref{sec:method}~and~\ref{sec:experiments}.

\section{\modelname -  Consistency and Histogram Optimization}
\label{sec:method}

\textbf{Preliminaries.} We are given a graph $\mathcal{G}=(\mathcal{V}, A)$ where $\mathcal{V}$ is a set of nodes $\{v_1,...,v_n\}$ and $A$ is the adjacency matrix, i.e,
\[A_{ij}=\begin{cases}
1 & v_{i}\ \mathrm{is\ directly\ connected\ to\ }v_{j}\\
0 & \mathrm{else}
\end{cases}\]
We ignore self-loops in the graph, hence $A_{ii}=0$ for all $i\in\{1,...,n\}$. In addition, we are given node feature matrix $X\in\mathbb{R}^{n\times d}$ where its $i$'th row is the feature vector of node $v_i$ and denoted by $\textbf{x}_i$. We define the train set $\mathcal{V}_{train}\subset\mathcal{V}$ and the validation set $\mathcal{V}_{val}\subset\mathcal{V}$. For each $v_i\in\mathcal{V}_{train}\cup\mathcal{V}_{val}$ we are given a label $\textbf{y}_i\in\{0,1\}^C$ encoded as a one-hot vector, where $C$ is the number of classes. We denote by \(d\left(u,v\right)\) the length of the shortest path in \(\mathcal{G}\) between the nodes $u$ and $v$. Furthermore, we denote the set of nodes that can be reached from $v$ with paths of distance no longer than $\ell$ by $\mathcal{N}^\ell\left(v\right)$, i.e, $\mathcal{N}^\ell\left(v\right)=\{u\in\mathcal{V}|d(v,u)\leq\ell\}$. We omit the superscript for $\ell=1$, denoting $\mathcal{N}^1\left(v\right)$ as $\mathcal{N}\left(v\right)$.

Our goal is to predict the labels of all the nodes in $\mathcal{V}/(\mathcal{V}_{train}\cup\mathcal{V}_{val})$. Following common practices, we only use $\mathcal{V}_{train}$ for optimizing the model weights and $\mathcal{V}_{val}$ for hyper-parameter selection. Note that we described the transductive settings, where all the nodes of the test-set are accessible during training. We describe the inductive settings, where some nodes of the test set are not present during training and the corresponding experiments in Sec. \ref{sec:inductive}.

\subsection{Prediction Network}

\begin{table}[!t]
\centering
\vskip -0.1in
\caption{Notation summary}
\vskip 0.05in
\begin{tabular*}{\linewidth}{ll}
\hline
{\textbf{Notation}} & {\textbf{Explanation}} \\\hline
$\mathcal{G}$  & Graph    \\
$\mathcal{V}$  & Set of nodes \\
$\mathcal{V}_{train}$  & Train set, subset of $\mathcal{V}$ \\
$\mathcal{V}_{val}$  & Validation set, subset of $\mathcal{V}$ \\
$A$            & Adjacency matrix\\
$d(v,u)$       & Length of the shortest path between $u$ and $v$\\
$\mathcal{N}\left(v\right)$  & Set of immediate neighbors of $v$\\
$\mathcal{N}^\ell\left(v\right)$  & Set of nodes $u \in \mathcal{V}$ s.t. $d(v,u)\leq \ell$\\
$X$            & Feature matrix in $\mathbb{R}^{n \times d}$ \\
$\mathbf{x}_i$ & The $i$'th row of $X$\\
$C$            & Number of classes\\
$\mathbf{y}_i$ & One-hot label of $v_i$ in $\{0,1\}^C$\\
\hline
\end{tabular*}
\label{NOTATION_TABLE} 
\vskip -0.1in
\end{table}

Our method uses a linear model as a backbone on all the datasets, except for the OGB datasets (\textit{ogbn-arxiv} and \textit{ogbn-products}) where we used two-layer MLPs. We denote the predictor as $\Psi$. The predictor is trained using the standard cross-entropy loss, between the prediction of the model and the provided ground truth (GT) labels. Formally, the loss is given by:
\begin{equation}\label{eq:gt_loss}
\mathcal{L}_{GT}\left(\Psi\right)=\sum_{\left(\textbf{x}_{i},\textbf{y}_{i}\right)\in\mathcal{V}_{train}}CE\left(\Psi\left(\textbf{x}_{i}\right),\textbf{y}_{i}\right)
\end{equation}

\subsection{Consistency Loss}\label{sec:consistency}

We showed in Sec.~\ref{sec:motivation} that training a simple node-level classifier tends to overfit on standard node-classification datasets due to very limited training set sizes. To overcome this limitation, we propose to incorporate powerful regularization loss utilizing graph priors. Concretely, we incorporate a homophilic prior on the node predictions using a consistency loss. In many node classification tasks, such as predicting attributes of academic papers in a citation network or attributes of products in a co-purchase network, neighboring nodes often have the same label. Consequently, homophilic priors enforce label consistency between neighboring nodes. In practice, our model does not output a single label but rather a probability distribution over the classes for each node. Therefore in order to enforce consistency we use a probability discrepancy measure between the predictions of adjacent nodes.
Specifically, we compute the average cross-entropy between predicted label distribution for a node and each of its neighbors. This loss term tends to produce consistent predictions for adjacent nodes.

We formulate the consistency loss term as:
\begin{equation}
\mathcal{L}_{consist}(\Psi)=
\small{\sum_{v_i\in \mathcal{V}}\left(\frac{1}{\left|\mathcal{N}\left(v_i\right)\right|}\sum_{v_j\in\mathcal{N}\left(v_i\right)}CE\left(\Psi\left(\textbf{x}_i\right),\Psi\left(\textbf{x}_j\right)\right)\right)}
\end{equation}

Our complete loss function is:
\begin{equation}\label{eq:loss}
\mathcal{L}\left(\Psi\right)=\mathcal{L}_{GT}\left(\Psi\right)+\gamma\cdot\mathcal{L}_{consist}\left(\Psi\right)
\end{equation}
Where $\gamma$ is a hyper-parameter controlling consistency regularization strength.

\subsection{Pseudo-Labelling Iterations}\label{sec:PseudoLabelling}

\begin{algorithm}[t]
  \caption{Pseudo-code of \modelname}
  \begin{algorithmic} %
  \STATE \textbf{Input:} Graph $\mathcal{G}$, Train-set $\mathcal{V}_{train}$, Iterations \emph{T}, Confidence threshold $\tau$
  \STATE Compute histograms according to Eq. \ref{eq:histograms} into $H$
  \STATE $X\leftarrow \texttt{Concatenate}(X,H)$
  \STATE Initialize $\Psi$, an MLP or linear model.
  \STATE \(\mathcal{V}^1_{train} = \mathcal{V}_{train}\)
  \FOR{\emph{t} = 1 to \emph{T}}
    \STATE Train $\Psi$ on $\mathcal{V}^t_{train}$ and $\mathcal{G}$ using the loss from Eq. \ref{eq:loss}
    \STATE \( \hat{Y} = \Psi(X)\)
    \STATE \(Y^* = \lambda \cdot \hat{Y}  + (1 - \lambda) \cdot \hat{A}\hat{Y}\)
    \STATE $\mathcal{V}^{t+1}_{train} \leftarrow \mathcal{V}_{train}\cup\{i|\max_{j=1,...,C}(Y^*_{ij}) > \tau\}$
  \ENDFOR
\end{algorithmic}
\end{algorithm}

\begin{table*}[thb!]
\caption{\modelname  achieves better average accuracy than graph-distillation methods and a popular GNN model. Results show accuracy in the transductive settings (higher is better).} \label{tab:maintable}
\setlength\tabcolsep{0pt} %
\footnotesize\centering
\smallskip 
\begin{tabular*}{\textwidth}{c@{\extracolsep{\fill}}lccccc}
\toprule
&  Dataset  & SAGE & GLNN & NOSMOG & \modelname  \\
\midrule
&\cora      & 80.52 $\pm$ 1.77  & 80.54 $\pm$ 1.35 & 83.04 $\pm$ 1.26 & 82.92 $\pm$ 1.15\\ 
&\citseer   & 70.33 $\pm$ 1.97  & 71.77 $\pm$ 2.01 & 73.78 $\pm$ 1.54 & 75.64 $\pm$ 1.68\\
&\pubmed    & 75.39 $\pm$ 2.09  & 75.42 $\pm$ 2.31 & 77.34 $\pm$ 2.36 & 77.22 $\pm$ 2.49\\ 
&\acomputer & 82.97 $\pm$ 2.16  & 83.03 $\pm$ 1.87 & 84.04 $\pm$ 1.01 & 81.03 $\pm$ 1.60\\
&\aphotos   & 90.90 $\pm$ 0.84  & 92.11 $\pm$ 1.08 & 93.36 $\pm$ 0.69 & 93.06 $\pm$ 1.56\\
&\ogba      & 70.92 $\pm$ 0.17  & 72.15 $\pm$ 0.27 & 71.65 $\pm$ 0.29 & 71.35 $\pm$ 0.25\\
&\ogbp      & 78.61 $\pm$ 0.49  & 77.65 $\pm$ 0.48 & 78.45 $\pm$ 0.38 & 81.71 $\pm$ 0.26\\
\midrule
&  Mean     & 78.52 & 78.95 & 80.24 & \textbf{80.42} \\
\bottomrule
\end{tabular*}
\end{table*}

Most standard node classification benchmarks are effectively semi-supervised. I.e., their training set is very small (sometimes as small as 0.3\% of the total number of nodes). To have a larger effective training set for our classifier, we take the predicted labels for some unlabelled nodes and add them to the training set. 

Specifically, before each iteration, we add to the training set the nodes on which the model made prediction with high confidence along with the existing ground truth training nodes. We take the model prediction as the labels of these non-ground-truth nodes and refer to them as \textit{pseudo-labels}. This set is used for training the prediction network in the following iteration. The ground-truth training nodes remain in the training set in all iterations, but the pseudo-labeled high-confidence nodes are recomputed for each iteration. I.e. if a high-confidence node becomes a low-confidence node in the following iteration, we will exclude it from the training set, unless its ground-truth label was provided. This adaptive mechanism allows the model to rectify erroneous early decisions as training progresses. The training set that we use in iteration $I+1$ is given by the following rule:

\begin{equation}
\mathcal{V}_{train}^{I+1}=\mathcal{V}_{train}\cup\{v_{i}|\max_{j=1,...,C}\left(\Psi^{I}\left(\textbf{x}_{i}\right)\right)_{j}>\tau\}
\end{equation}
Where $\Psi^{I}$ is the model that was trained on the training set $\mathcal{V}_{train}^{I}$, and $\tau$ is the confidence threshold.

\textbf{Pseudo-Label Smoothing.} We found that on many popular datasets, smoothing the model predictions on the target node with the predictions on the neighboring nodes can result in higher performance. Unfortunately, this improvement comes at the cost of increased computational complexity during inference, as it requires evaluating the model on neighboring nodes. To address this challenge, we apply prediction smoothing only to the pseudo-labels within the iterative training algorithm (but not at inference time). We found that with this strategy, our model achieves similar final accuracy to that achieved through test-time smoothing, without actually smoothing at test time. This observation suggests that the model learns to integrate the homophilic prior into its predictions.

The smoothing technique applied to pseudo-labels involves generating predictions \(\hat{Y} \in \mathbb{R}^{n \times C}\) for all nodes after each training iteration. Each row of \(\hat{Y}\) represents the predicted distribution vector for a specific node. Subsequently, an adjusted prediction \(Y^*\) is computed for each node by taking a weighted average between its own prediction and the average prediction of its neighbors. The weighting factor \(\lambda\), determined empirically using the validation set, is introduced in the smoothing process through the equation:
\begin{equation}
   Y^* = \lambda \cdot \hat{Y}  + (1 - \lambda) \cdot \hat{A}\hat{Y}
\end{equation}
Here, \(\hat{A}\) denotes the normalized adjacency matrix, such that the \(i\)-th row of \(\hat{A}\hat{Y}\) represents the average prediction of the neighbors of node \(i\). The resulting \(Y^*\) serves as the refined prediction used in the pseudo-labeling strategy described above.

\subsection{Histograms of Neighboring Labels}\label{sec:histograms}

The neighborhood of a node may allow us to infer further information about that specific node label (which is not as relevant to further nodes further away). Local information has been incorporated into previous methods in various ways. For example, NOSMOG \cite{nosmogcite} incorporates a positional embedding, DeepWalk \cite{deepwalk}, in its features. This embedding allows the classifier to learn a connection between a node position in the graph and the labels of the nodes around it. Here, we incorporate the information on the labels of neighbors directly.

We propose a method for using a descriptor of neighbor labels to improve our predictor. Our approach differs from GNNs, as it only aggregates the provided labels from neighboring nodes. On the other hand, GNNs use message passing, which requires computing hidden features for the entire neighborhood. Our method only requires simple counting of the neighborhood labels which is a much weaker requirement than running a GNN over the entire neighborhood.

Specifically, for each node \(v\), the descriptor is a weighted histogram derived from the labels of all nodes with a path length to \(v\) not exceeding \(\ell\). The weight assigned to each node in the histogram is determined by the distance between the respective node and \(v\) - that is, the minimal length of a path between them. 

The histogram descriptor \(\textbf{h}_i\) for a node \(v_i\) is calculated by first computing  $\textbf{h}'_{i}$ as follows:
\begin{equation}\label{eq:exact_histogram}
\textbf{h}'_{i}=\sum_{v_{j}\in\mathcal{N}^{\ell}\left(v_{i}\right)\cap\mathcal{V}_{train}}\left(\alpha^{d\left(v_{i},v_{j}\right)}\cdot\textbf{y }_{j}\right)
\end{equation}
Here, $\alpha \in [0,1]$ is a hyper-parameter controlling the relative importance of far away nodes. Since $\textbf{y }_j$ is a one-hot vector in $\{0,1\}^C$, $\textbf{h}'_i$ represents a weighted sum of labels from nodes within a local context of \(v_i\), with the size of the context determined by \(\ell\).

Subsequently, to obtain a normalized histogram,  $\textbf{h}_i$, we divide $\textbf{h}'_i$ by its sum, this descriptor is concatenated to the original input vector $\textbf{x}_i$.
\begin{equation}\label{eq:histograms}
\textbf{h}_{i}=\frac{\textbf{h}'_{i}}{\sum_{j=1}^{C}\textbf{h}'_{ij}}
\end{equation}

The requirement of determining the distance between each node in the training set and all other nodes in the graph, is a task with a computational complexity of $\mathcal{O}(\left|\mathcal{V}_{train}\right|\cdot\left|E\right|)$, where $\left|E\right|$ is the number of edges in the graph (assuming there are more edges than nodes). In the standard setting for node-level classification tasks, the size of \(\mathcal{V}_{train}\) is often very small, so computing the histograms is feasible.

Yet, for larger datasets, such as \textit{ogbn-products}, this calculation becomes cumbersome. To address this, we propose an efficient approximation for $\textbf{h}'_{i}$. We calculate histogram for all nodes in the graph jointly by spreading the labels using convolution operations. Specifically, the matrix $H'$ whose rows represent un-normalized histograms for each node, is obtained by:
\begin{equation}\label{eq:approx_histograms}
H'=\sum_{k=1}^{\ell}\left(\alpha\hat{A}\right)^{k}\Tilde{Y}
\end{equation}
Where the $i$'th row of $\Tilde{Y}$ is defined by:
\[
\Tilde{\textbf{y}}_i=\begin{cases}
\textbf{y}_{i} & v_{i}\in\mathcal{V}_{train}\\
\textbf{0} & v_{i}\notin\mathcal{V}_{train}
\end{cases}
\]
We normalize \(H'\) in the same manner presented in Eq. \ref{eq:histograms}.

When using convolutions, this computation takes a running time of $\mathcal{O}(\left|E\right|)$. However,
unlike the previous method of calculating histograms, the labels of some nodes in the training might leak into the label histogram feature. This can affect the generalization, as we do not have this information at test time. We observed that using small enough values for $\alpha$ eliminated the generalization gap due to this issue.

\begin{table*}
\caption{Ablation table}\label{tab:Ablation2}
\vskip 0.1in
\centering
    \begin{tabular*}{\textwidth}{@{\extracolsep{\fill}}lccccc}
        \toprule
        Dataset & \modelname & \makecell{Only Iterative Training}& \makecell{Only Histograms}& \makecell{Only Consistency Loss}& Base \\
        \midrule
        \texttt{cora}        & 82.92 & 78.39 & 78.86 & 73.59 & 65.26 \\
        \texttt{citeseer} & 75.64 & 73.97 & 73.02 & 73.38 & 69.43 \\
        \texttt{pubmed} & 77.22 & 71.98 & 75.77 & 68.86 & 68.8 \\
        \texttt{a-computer} & 81.03 & 78.75 & 76.45 & 72.1 & 72.68 \\
        \texttt{a-photo} & 93.06 & 89.19 & 86.59 & 86.43 & 83.11 \\
        \midrule
        \textbf{Mean} & \textbf{81.97} & \textbf{78.5} & \textbf{78.1} & \textbf{74.9} & \textbf{71.86} \\
        \bottomrule
    \end{tabular*}
\end{table*}

\section{Experiments}\label{sec:experiments}

We empirically validate our approach on seven publicly available graph benchmark datasets, and compare it to state-of-the-art graph-distillation based methods. We then ablate the different components of our methods and evaluate the conditions under which they are significant.

\subsection{Datasets}
We evaluate a selection of datasets commonly used in the graph learning community.
We follow previous works \cite{yang2021extract,glnncite, nosmogcite} in only considering the largest connected component of each graph dataset, and regard the edges as undirected. The statistics of all the datasets are presented in App. Tab.~\ref{tab:dataset}.

\textit{\textbf{Cora}}, \textit{\textbf{CiteSeer}} \cite{coraciteseer} and \textit{\textbf{PubMed}} \cite{pubmed} are citation networks where each node represents a scientific paper, edges signify citations between papers, and labels denote the research field of each paper. In \textit{Cora} and \textit{CiteSeer} the feature vector of each node is a sparse bag-of-words derived from the text of the paper. \textit{PubMed} is constructed from medical publications, the node features are represented by TF/IDF \cite{tfidf} weighted word frequency. The labels indicate the type of diabetes the publication focuses on.

\textit{\textbf{A-Computers}} and \textit{\textbf{A-Photo}} \cite{amazondatasets} are extracted from the Amazon co-purchase graph \cite{amazongraph}. These datasets involve nodes representing electronic goods sold on Amazon web store. Edges indicate whether two products are frequently bought together. The node features are product reviews encoded using a bag-of-words representation. The labels assigned to the nodes correspond to product categories, with \textit{A-Computers} encompassing categories such as Desktops, Laptops, Monitors, and so forth. \textit{A-Photo} includes categories such as Cameras, Lenses etc.

\textbf{\textit{ogbn-arxiv}} and \textbf{\textit{ogbn-products}} are from the Open Graph Benchmark (OGB) \cite{ogbndatasets} and are larger datasets. The former constitutes a citation network of arXiv papers, while the latter is a co-purchasing network. 

\textbf{Dataset split.} We follow the protocol used in previous studies for partitioning datasets into training, validation, and test sets. In the transductive setting, \textit{Cora}, \textit{CiteSeer}, \textit{PubMed}, \textit{A-Computers} and \textit{A-Photo} are partitioned by randomly sampling 20 instances per class for training, 30 instances per class for validation, and treating the remaining nodes as the test set. For the \textit{ogbn-arxiv} dataset, the training is conducted on papers published until 2017, validation is performed on those published in 2018, and testing is carried out on papers published since 2019. In the case of \textit{ogbn-products}, nodes (representing products) are arranged based on their sales ranking. The top 8\% of products are assigned to the training set, the subsequent top 2\% to the validation set, and the remaining products constitute the test set. The partitioning scheme used by the OGB datasets is designed to perform an accurate simulation of real-life scenarios.

\textbf{Baselines.} We compare our method to recent 
GNN distillation-based methods: (1) GLNN - A graph learning based method which uses knowledge distillation from a pre-trained GNN teacher model to an MLP student model. The student is trained to predict the soft-labels obtained from the teacher. (2) NOSMOG - In addition to training an MLP on soft-labels, this method adds an adversarial feature augmentation loss, similarity distillation of hidden features and fusing positional encoding features to the input. We further compare to the teacher used in the KD methods - GraphSAGE  with GCN aggregation strategy.

\subsection{Results}
Our approach yields competitive results despite not requiring the training of any graph neural networks. As shown in table \ref{tab:maintable}, our method achieves better accuracy, on average, across all the seven datasets, compared to the baselines. The performance of $\modelname$ on citation network datasets is particularly noteworthy, where it achieves markedly improved results on specific datasets. This observation holds true for both standard transductive settings and the more challenging inductive settings, as discussed in Section \ref{sec:inductive}. While our method achieved only slightly better accuracy than NOSMOG, it does so without training any GNN. This result allows us to obtain interesting insights into the required information; learning a simple model that performs well on complex graph structured data.

CoHOp achieves strong results on the larger datasets obtained from the Open Graph Benchmark (OGB) without iterative training. This is primarily because these datasets include large training splits, both in terms of the proportion of the entire graph which carries labels at train time, and in terms of absolute number of labeled nodes. Consequently, in these cases, we chose to include only the consistency loss and label-histogram feature augmentations as part of our method.

\subsubsection{Inductive Setting}\label{sec:inductive}

In the inductive setting we further split the unlabelled test set, denoted as $\mathcal{U}=\mathcal{V}/(\mathcal{V}_{train}\cup\mathcal{V}_{val})$, into two disjoint sets: (1) Unseen test nodes, a set of nodes exclusively available during inference time and not in training time, denoted by $\mathcal{U}_{unseen}$. (2) Observed test nodes, a set of unlabeled nodes  with accessible features during training, denoted by $\mathcal{U}_{seen}$. Unlike the unseen test nodes, the observed test nodes participate in the consistency loss and may have pseudo-labels. In the inductive setting, we train our model on the graph induced by all the nodes in the set $\mathcal{V}_{train}\cup\mathcal{V}_{val}\cup\mathcal{U}_{seen}$. Only the nodes of $\mathcal{V}_{train}$ are used in the classification loss (Eq. \ref{eq:gt_loss}). At training time, we discard edges that connect to nodes that are in $\mathcal{U}_{unseen}$. The test accuracy is computed on the combination of the sets $\mathcal{U}_{seen}$ and $\mathcal{U}_{unseen}$.

Similarly to the transductive case, our approach provides better accuracy on average across all the datasets, as shown in App.Tab. \ref{tab:inductive}.

\subsection{Ablation study}

CoHOp integrates diverse techniques aimed at enhancing a simple predictor by using the graph structure. These include: iterative training to address the limitations of an overly limited training set, a consistency loss for controlling the smoothness of the model's predictions, and feature augmentation with label histograms. This section examines the impact of these techniques on our overall approach.

We start by examining our classifier with only the standard cross-entropy classification loss. We see from Tab. \ref{tab:Ablation2} that the results of this naive classifier are far from being competitive. We show that each of the proposed components has a significant impact on the performance. As can be seen in App.Tab. \ref{tab:Ablation1}, pairs of these components already achieve strong results. Yet, the best performance is achieved when using our full method.

\subsubsection{Homophily Prior}

\begin{table}[!t]
\centering
\vskip -0.1in
\caption{Incorporating the homophily prior, through consistency loss and pseudo-labels smoothing, improves the model's accuracy by 2.4\% on average.}
\label{tab:homophily_prior}
\vskip 0.1in
\begin{tabular*}{\linewidth}{c@{\extracolsep{\fill}}lccc}
    \toprule
    & Dataset    & W.o. Homophily Prior & $\Delta$ \\
    \midrule
    & \texttt{cora}       & 79.64 & -3.28 \\
    & \texttt{citeseer}   & 73.97 & -1.67 \\
    & \texttt{pubmed}     & 76.61 & -0.61 \\
    & \texttt{a-computer} & 79.21 & -1.82 \\
    & \texttt{a-photo}    & 87.44 & -5.62 \\
    & \ogba$^*$           & 71.35 & 0 \\
    & \ogbp$^*$           & 78.07 & -3.64 \\
    \midrule
    & \textbf{Mean} & \textbf{78.04}& \textbf{-2.38}\\
    \bottomrule
\end{tabular*}
\begin{flushleft}\small 
$^*$Since we do not use iterative training in these datasets, accuracy only reflects the impact of the consistency loss.
\end{flushleft}
\vskip -0.2in
\end{table}

The use of the homophily prior, which posits that neighboring nodes exhibit positively correlated labels, is reflected in our method in two ways: (1) The inclusion of a consistency loss, which encourages the model to maintain correlation among neighboring nodes. (2) Smoothing the pseudo-labels used during iterative training, further incentivizing
the model to provide smoothed predictions across graph edges. In this section, we study the advantages associated with incorporating this homophily prior.

As depicted in Table \ref{tab:homophily_prior}, incorporating these two elements in our method resulted in an average improvement of 2.4\% across all the datasets including in this ablation study. This is explained by the homophilic tendencies of the dataset sources. For instance, in citation networks, it is reasonable to expect that papers in the same field may cite each other. Similarly, in co-purchasing networks, it is plausible that customers tend to buy items from the same category at the same time.

The \textit{ogbn-arxiv} dataset is an outlier as it is not positively affected by the homophily prior. Notably, this dataset has the largest proportion (53.7\%) of labeled training nodes. As the number of labels is sufficient, the advantage of using additional priors such as homophily decreases significantly.

\subsubsection{Histogram Approximation}

\begin{figure}[!t]
  \centering
  \begin{tikzpicture}
    \begin{axis}[
      ybar,
      width=\linewidth,
      height=5cm,
      grid=major,
      nodes near coords,
      nodes near coords align={vertical},
      nodes near coords style={font=\scriptsize\sansmath\sffamily},
      bar width=0.45cm,
      legend style={at={(0.5,-0.15)}, anchor=north, legend columns=-1},
      tick label style={font=\sansmath\sffamily\small},
      axis x line=bottom,
      symbolic x coords={cora, citeseer, pubmed, computer, photo},
      xtick=data,
      enlarge x limits=0.1,
      xticklabel style={font=\scriptsize\sansmath\sffamily, text height=0.7ex},
      xmajorgrids=false,
      axis y line=left,
      ylabel={Accuracy (\%)},
      ymin=0,
      ymax=100
    ]
    
    \addplot[fill=blue!50] coordinates {
      (cora, 83)
      (citeseer, 75.6)
      (pubmed, 77.2)
      (computer, 81)
      (photo, 93)
    };
    
    \addplot[fill=red!50] coordinates {
      (cora, 80.8)
      (citeseer, 73.6)
      (pubmed, 71.1)
      (computer, 82.3)
      (photo, 93.1)
    };

    \legend{Exact, Approx.}
    \end{axis}
  \end{tikzpicture}
  \vspace{-10pt}
  \caption{ 
The model's accuracy is 1.66\% higher on average when using the exact formula (Eq. \ref{eq:exact_histogram}) for histogram calculation compared to its approximation (Eq. \ref{eq:approx_histograms}).}
\label{fig:histograms_acc}
\end{figure}
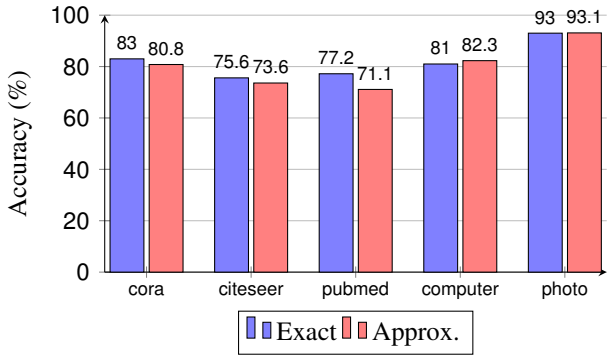

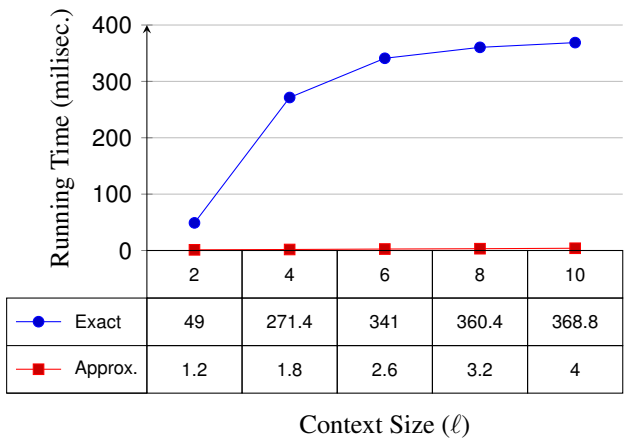
\begin{figure}[!t]
    \pgfplotsset{compat=1.14}
\pgfplotstableread{
x a b
2 49	1.2
4 271.4	1.8
6 341	2.6
8 360.4	3.2
10 368.8	4

}\datatable

\pgfplotstablegetrowsof{\datatable}
\pgfmathsetmacro{\Nrows}{\pgfplotsretval-1}
\pgfmathsetlengthmacro{\MyAxisW}{40ex}

\begin{tikzpicture}[
  cell/.style={ %
    draw,
    minimum width={\MyAxisW/(\Nrows+1)}, %
    minimum height=4ex,
    inner sep=0pt,
    outer sep=0pt,
    anchor=north west,
    font=\sffamily\scriptsize,
  }]
\begin{axis}[
   name=ax,
   scale only axis,
   width=\MyAxisW,
   height=3cm,
   xtick=\empty,
   grid=major,
   axis y line=left,
   x axis line style={draw=none},
   enlarge x limits={abs={\MyAxisW/(2*\Nrows+2)}},
   ymin=0,
   ymax=400,
   title style={font=\bfseries},
   ylabel={Running Time (milisec.)},
   xlabel={Context Size ($\ell$)},
   xlabel shift=13ex,
   tick label style={font=\sansmath\sffamily\small},
  ]
  \addplot table[x expr=\coordindex,y=a] {\datatable};
  \label{dataA}

  \addplot table[x expr=\coordindex,y=b] {\datatable};
  \label{dataB}
\end{axis}

\coordinate (c-0-0) at (ax.south west);

\foreach [count=\j from 1] \i in {0,...,\Nrows}
  {
  \pgfplotstablegetelem{\i}{x}\of\datatable
  \node [cell] (c-0-\j) at (c-0-\i.north east) {\pgfplotsretval};
  \pgfplotstablegetelem{\i}{a}\of\datatable
  \node [cell] (c-1-\j) at (c-0-\j.south west) {\pgfplotsretval};
  \pgfplotstablegetelem{\i}{b}\of\datatable
  \node [cell] (c-2-\j) at (c-1-\j.south west) {\pgfplotsretval};
  }

\matrix [draw,nodes={cell,minimum width=0pt,draw=none},anchor=north east,row sep=0pt,column sep=5pt,outer sep=0pt,inner ysep=0pt] (m) at (c-1-1.north west)
{
 \node {\ref{dataA}};  & \node{Exact}; \\
 \node {\ref{dataB}}; & \node{Approx.}; \\
};

\draw (m.west |- c-1-1.south west) -- (c-1-1.south west);
\draw (m.west |- c-2-1.south west) -- (c-2-1.south west);

\end{tikzpicture}
    \vspace{-25pt} %
    \caption{Running times of the pre-processing procedure of calculating the histograms using the exact calculation (Eq. \ref{eq:exact_histogram}) and using its approximation (Eq. \ref{eq:approx_histograms}) as function as the context size.}
    \label{fig:histograms_runtime}
     \vspace{-10pt}
\end{figure}

As described in Sec. \ref{sec:histograms}, we augment the input features by concatenating histograms of labels  derived from the local context of each node. In large datasets, we speed the method up by approximating the histograms using convolutions. In this section we analyze the trade-off between the time saved by the approximation and its potential impact on accuracy compared to exact computation. The results are presented in Fig.~\ref{fig:histograms_acc} and Fig.~\ref{fig:histograms_runtime}. The results show that while the exact histogram calculation results in a superior accuracy of 1.66\% on average, approximating the histograms significantly reduces computation time.

\section{Discussion}

\textbf{Utilization of unlabelled data.} Our analysis of distillation methods revealed that their main success factor is efficiently using the unlabeled data (see Sec. \ref{sec:motivation}). However, to do so these methods needed to train GNNs and distill them on the unlabeled nodes. In contrast, our method was able to bypass training a GNN entirely using the proposed iterative training scheme. 

\textbf{Regularization.} Our method further exploits the unlabeled nodes by integrating a consistency loss that considers all the nodes visible at training time. Distillation methods enforce smoothness only indirectly through the distillation of GNNs, which tend to produce similar predictions for neighboring nodes as they aggregate features across adjacent nodes \cite{gnnoversmoothing}. This consistency regularization proves crucial in mitigating overfitting challenges encountered when training MLPs on datasets that have very few training labels. Interestingly, it also plays an important role on datasets that had many training labels but characterized by a distributional mismatch between labeled training nodes and unlabeled test nodes.

\textbf{Augmenting node features.} In many graph datasets, the labels of neighboring nodes are highly informative for node classification. In the common homophilic case, the correlation between the labels of neighboring nodes is highly positive. Our method benefits from this prior by augmenting node features with the label-histograms of the neighboring nodes. Some distillation methods, such as NOSMOG, augment node features using positional embeddings, particularly DeepWalk. However, our label-histograms are much more computationally efficient than DeepWalk positional features.

\section{Limitations}

\textbf{Heterophilic graphs.} Our method is focused on utilizing the homophilic prior. Yet, some graph datasets \cite{hetrographs} are heterophilic. In such datasets, while the label of each node does not tend to be similar to the label of its neighbors, labels of neighboring nodes may still carry information. Some components of our method may be suitable for heterophilic graphs. E.g., neighborhood histograms may carry information about the labels of the node, even when neighbors tend to come from different classes. Adapting our method to heterophilic graphs is left for future work.

\textbf{Dataset-specific variability.} While our method outperforms NOSMOG on average, there are cases where NOSMOG achieves higher accuracy. This variation suggests that the different approaches might be influenced by dataset-specific characteristics. Further investigation into specific cases where NOSMOG outperforms may allow future research to develop methods that enjoy the best of all worlds.

\section{Conclusion}
We introduced \modelname, a fully GNN-free method designed for node classification. We initially established that simple node classifiers, without distillation, can match the performance of GNNs or distillation methods on benchmark datasets, if the number of labeled nodes is sufficiently large. The core challenge we tackled was adapting GNN-less MLP methods to the label-poor setting popular in node classification benchmarks. Our proposed method consisted of three core components: a label consistency loss, an iterative labeling scheme, and feature augmentation using label histograms.
Our method was able to surpass or match node classification accuracy on multiple popular benchmark datasets.

\newpage

\section{Acknowledgment}
This research was partially supported by the Israeli data science scholarship for outstanding postdoctoral fellows (VATAT).

\bibliography{references}
\bibliographystyle{conference_bib_style.bst}

\newpage
\appendix
\onecolumn

\section{Appendix}

\subsection{Implementation details}

\begin{itemize}
    \item We conducted each experiment using 10 different random seeds and reported the mean and standard deviation of the model accuracy on the test set.
    
    \item The backbone we employed consists of a single linear layer for all datasets, except for OGB datasets, where we utilized a two-layer MLP with hidden dimensions of 512 and 1024 for \textit{ogbn-products} and \textit{ogbn-arxiv}, respectively.
    
    \item Across all datasets, except for \textit{ogbn-products} and \textit{ogbn-arxiv}, we employed 5 iterations, as outlined in Section \ref{sec:PseudoLabelling}. Within each iteration, the model underwent training for 200 epochs, and the optimal epoch was determined based on performance on the validation set. Notably, for the \textit{ogbn-products} dataset, it was observed that a single epoch and iteration sufficed, owing to the dataset's substantial size.
    
    \item The parameter defining the size of the local context utilized for computing the histogram, as explained in Section \ref{sec:histograms}, was set to 10 hops (i.e., $\ell=10$).

    \item We used Equation \ref{eq:approx_histograms} for calculation the approximation of the histograms in the larger datasets from OGB. While for all the other datasets we used the original formula (Eq. \ref{eq:exact_histogram}).
     \item The weights of the model are initialized once, then at each iteration we continue the training for numerous epochs.
\end{itemize}

\textbf{Python Libraries.} We use Deep Graph Library (DGL) \cite{dgl} for storing the graph datasets and performing graph operations on them. We also use PyTorch \cite{pytorch} and scikit-learn \cite{scikitlearn}.

\begin{table*}[ht]
\center
\caption{Dataset Statistics.}
\vskip 0.1in 
\begin{tabular*}{\textwidth}{@{}lllllll@{}}
\toprule
\multicolumn{1}{l}{Dataset} & \multicolumn{1}{l}{\# Nodes} & \# Edges & \# Features & \# Classes & Split Strategy & Split Sizes (train / val / test) \\
\midrule
\cora      & 2,485     & 5,069      & 1,433 & 7  & Random & 140 / 210 / 2,135  \\
\citseer   & 2,110     & 3,668      & 3,703 & 6  & Random & 120 / 180 / 1,810  \\ 
\pubmed    & 19,717    & 44,324     & 500   & 3  & Random & 60\,\,\,  / 90\,\,\,  / 19,567   \\ 
\acomputer & 13,381    & 245,778    & 767   & 10 & Random & 200 / 300 / 12,881 \\
\aphotos   & 7,487     & 119,043    & 745   & 8  & Random & 160 / 240 / 7,087  \\
\ogba      & 169,343   & 1,166,243  & 128   & 40 & Public & 53.7\% / 17.6\% / 28.7\%   \\
\ogbp      & 2,449,029 & 61,859,140 & 100   & 47 & Public & 8\% / 1.6\% / 90.4\%  \\ 
\bottomrule
\end{tabular*}
\label{tab:dataset}
\end{table*}

\begin{table*}[thb]
\caption{Ablation table}\label{tab:Ablation1}
\vskip 0.1in 
\centering
    \begin{tabular*}{\textwidth}{@{\extracolsep{\fill}}lccccc}
        \toprule
        Dataset & \modelname & \makecell{With Histograms \\ and Iterations}  & \makecell{With Consistency \\and Iterations} & \makecell{With Consistency \\ and Histograms} & \makecell{Base} \\
        \midrule
        \texttt{cora}        & 82.92 & 81.79 & 80.41 & 81.75 & 65.26 \\
        \texttt{citeseer} & 75.64 & 74.95 & 74.85 & 74.19 & 69.43 \\
        \texttt{pubmed} & 77.22 & 77.04 & 72.44 & 76.57 & 68.8 \\
        \texttt{a-computer} & 81.03 & 80.57 & 79.87 & 75.38 & 72.68 \\
        \texttt{a-photo} & 93.06 & 91.31 & 91.14 & 90.21 & 83.11 \\
        \midrule
        \textbf{Mean} & \textbf{81.97} & \textbf{81.13} & \textbf{79.74} & \textbf{79.62} & \textbf{71.86} \\
        \bottomrule
    \end{tabular*}
\end{table*}

\begin{table*}[!ht]
    \center
    \caption{Inductive setting: The test set is further partitioned into 80\% test set that present during training (\seen) and 20\% unseen test (\unseen). The formula used for computing a composite measure denoted as $prod$ is expressed as follows: \(prod = 0.8 \cdot seen + 0.2 \cdot unseen\).}
    \label{tab:inductive}
    \begin{adjustbox}{width=\textwidth,center}
\renewcommand{\arraystretch}{1.5}
\begin{tabular}{lllllllll}
\toprule

{Datasets} & Eval & SAGE & MLP & GLNN & \nosmog{}  &\modelname& $\Delta_{GLNN}$& $\Delta_{NOSMOG
}$\\ \midrule

\multirow{3}{*}{\cora} 
        & \production & 79.53 & 59.18 & 78.28
& 81.02  &\textbf{81.11}& $\uparrow$ 2.83\%
& $\uparrow$ 0.09\%
\\
        & \unseen        & 81.03 $\pm$ 1.71 & 59.44 $\pm$ 3.36 & 73.82 $\pm$ 1.93
& 81.36 $\pm$ 1.53  &80.09 $\pm$ 2.29& $\uparrow$ 6.27\%
& $\downarrow$ -1.27\%
\\
        & \seen       & 79.16 $\pm$ 1.60 & 59.12 $\pm$ 1.49 & 79.39 $\pm$ 1.64
& 80.93 $\pm$ 1.65  &81.37 $\pm$ 1.74& $\uparrow$ 1.98\%
& $\uparrow$ 0.44\%
\\  \midrule
\multirow{3}{*}{\citseer}  
        & \production & 68.06 & 58.49 & 69.27
& 70.60  &\textbf{72.94}& $\uparrow$ 3.67\%
& $\uparrow$ 2.34\%
\\ 
        & \unseen        & 69.14 $\pm$ 2.99 & 59.31 $\pm$ 4.56 & 69.25 $\pm$ 2.25
& 70.30 $\pm$ 2.30  &71.77 $\pm$ 3.37& $\uparrow$ 2.52\%
& $\uparrow$ 1.47\%
\\
        & \seen       & 67.79 $\pm$ 2.80 & 58.29 $\pm$ 1.94 & 69.28 $\pm$ 3.12
& 70.67 $\pm$ 2.25  &73.23 $\pm$ 3.13& $\uparrow$ 3.95\%
& $\uparrow$ 2.56\%
\\ \midrule
\multirow{3}{*}{\pubmed}  
        & \production & 74.77 & 68.39 & 74.71
& \textbf{75.82}&74.51& $\downarrow$ -0.2\%
& $\downarrow$ -1.31\%
\\ 
        & \unseen        & 75.07 $\pm$ 2.89 & 68.28 $\pm$ 3.25 & 74.3 $\pm$ 2.61
& 75.87 $\pm$ 3.32  &74.84 $\pm$ 3.39& $\uparrow$ 0.54\%
& $\downarrow$ -1.03\%
\\
        & \seen       & 74.70 $\pm$ 2.33 & 68.42 $\pm$ 3.06 & 74.81 $\pm$ 2.39
& 75.80 $\pm$ 3.06  &74.43 $\pm$ 3.20& $\downarrow$ -0.38\%
& $\downarrow$ -1.37\%
\\ \midrule
\multirow{3}{*}{\acomputer}  
        & \production & 82.73 & 67.62 & 82.29
& \textbf{83.85}&80.66& $\downarrow$ -1.63\%
& $\downarrow$ -3.19\%
\\
        & \unseen        & 82.83 $\pm$ 1.51 & 67.69 $\pm$ 2.20 & 80.92 $\pm$ 1.36
& 84.36 $\pm$ 1.57  &80.19 $\pm$ 2.65& $\downarrow$ -0.73\%
& $\downarrow$ -4.17\%
\\
        & \seen       & 82.70 $\pm$ 1.34 & 67.60 $\pm$ 2.23 & 82.63 $\pm$ 1.4
& 83.72 $\pm$ 1.44  &80.78 $\pm$ 2.29& $\downarrow$ -1.85\%
& $\downarrow$ -2.94\%
\\ \midrule
\multirow{3}{*}{\aphotos}  
        & \production & 90.45 & 77.29 & 92.38
& \textbf{92.47}&92.11& $\downarrow$ -0.27\%
& $\downarrow$ -0.36\%
\\ 
        & \unseen        & 90.56 $\pm$ 1.47 & 77.44 $\pm$ 1.50 & 91.18 $\pm$ 0.81
& 92.61 $\pm$ 1.09  &91.75 $\pm$ 1.36& $\uparrow$ 0.57\%
& $\downarrow$ -0.86\%
\\
        & \seen       & 90.42 $\pm$ 0.68 & 77.25 $\pm$ 1.90 & 92.68 $\pm$ 0.56
& 92.44 $\pm$ 0.51  &92.20 $\pm$ 1.35& $\downarrow$ -0.48\%
& $\downarrow$ -0.24\%
\\ \midrule
\multirow{3}{*}{\ogba}  
        & \production & 70.69 & 55.35 & 65.09
& 70.90  &\textbf{71.32}& $\uparrow$ 6.23\%
& $\uparrow$ 0.42\%
\\ 
        & \unseen        & 70.69 $\pm$ 0.58 & 55.29 $\pm$ 0.63 & 60.48 $\pm$ 0.46
& 70.09 $\pm$ 0.55  &71.42 $\pm$ 0.34& $\uparrow$ 10.94\%
& $\uparrow$ 1.33\%
\\
        & \seen       & 70.69 $\pm$ 0.39 & 55.36 $\pm$ 0.34 & 71.46 $\pm$ 0.33
& 71.10 $\pm$ 0.34  &71.29 $\pm$ 0.22& $\downarrow$ -0.17\%
& $\uparrow$ 0.19\%
\\ \midrule
\multirow{3}{*}{\ogbp}  
        & \production & 76.93 & 60.02 & 75.77
& 77.33  &\textbf{81.28}& $\uparrow$ 5.51\%
& $\uparrow$ 3.95\%
\\ 
        & \unseen        & 77.23 $\pm$ 0.24 & 60.02 $\pm$ 0.09 & 75.16 $\pm$ 0.34
& 77.02 $\pm$ 0.19  &81.69 $\pm$ 0.19& $\uparrow$ 6.53\%
& $\uparrow$ 4.67\%
\\
        & \seen       & 76.86 $\pm$ 0.27 & 60.02 $\pm$ 0.11 & 75.92 $\pm$ 0.61
& 77.41 $\pm$ 0.21  &81.18 $\pm$ 0.18& $\uparrow$ 5.26\%
& $\uparrow$ 3.77\%
\\
\midrule\midrule
\textbf{Mean} &  \production & 77.59& 63.76& 76.83& 78.86& \textbf{79.13}& $\uparrow$ 2.31\%& $\uparrow$ 0.28\%\\
\bottomrule
\end{tabular}
\end{adjustbox}

\end{table*}

\end{document}